\documentclass[11pt,a4paper,table]{article}
\usepackage[hyperref]{conll-2019}
\usepackage{times}
\usepackage{latexsym}
\usepackage{times}
\usepackage{latexsym}
\usepackage{times}
\usepackage{latexsym}
\usepackage{float}
\usepackage{algorithm}
\usepackage{amsfonts}
\newcommand{\Z}{\mathbb{Z}}
\newcommand{\R}{\mathbb{R}}
\usepackage{changepage}
\usepackage{makecell}
\usepackage[noend]{algpseudocode}
\usepackage{sidecap}
\algnewcommand\algorithmicinput{\textbf{Input:}}
\algnewcommand\algorithmicoutput{\textbf{Output:}}
\algnewcommand\Input{\item[\algorithmicinput]}%
\algnewcommand\Output{\item[\algorithmicoutput]}%
\usepackage{graphicx,rotating,booktabs}
\usepackage[verbose]{placeins}
\usepackage[ngerman]{babel}
\usepackage{blindtext}
\usepackage{diagbox}
\usepackage{url}

\usepackage{amssymb}
\usepackage{pifont}
\usepackage{times}
\usepackage{latexsym}
\usepackage{times}
\usepackage{latexsym}
\usepackage{float}
\usepackage{algorithm}
\usepackage{amsfonts}
\usepackage{changepage}

\usepackage[noend]{algpseudocode}
\usepackage{sidecap}
\usepackage{graphicx,rotating,booktabs}
\usepackage[verbose]{placeins}
\usepackage[ngerman]{babel}
\usepackage{blindtext}
\usepackage{diagbox}
\usepackage{url}
\usepackage{makecell}

\usepackage{amssymb}
\usepackage{pifont}

\newcommand{\st}{Stress Test}
\newcommand{\rte}{RTE-Quant}
\newcommand{\awp}{AwpNLI}
\newcommand{\qnli}{NewsNLI}
\newcommand{\reddit}{RedditNLI}
\newcommand{\Qreas}{\textsc{Q-REAS} }
\newcommand{\numset}{\textsc{Numset} }
\newcommand{\numsets}{\textsc{Numsets} }
\usepackage{url}

\usepackage{url}

\aclfinalcopy 

\title{EQUATE : A Benchmark Evaluation Framework for Quantitative Reasoning in Natural Language Inference}

\author{Abhilasha Ravichander\thanks{*The first two authors contributed equally to this work.} , Aakanksha Naik\footnotemark[1] , \\   \textbf{Carolyn Rose}, \textbf{Eduard Hovy} \\
Language Technologies Institute, Carnegie Mellon University\\
  {\tt\{aravicha, anaik, cprose, hovy\}@cs.cmu.edu}}
\date{}

\date{}

\begin{document}
\maketitle
\begin{abstract}
Quantitative reasoning is a higher-order reasoning skill that any intelligent natural language understanding system can reasonably be expected to handle. We present \textsc{equate}\footnote{Code and data available at \url{https://github.com/AbhilashaRavichander/EQUATE}.} (Evaluating Quantitative Understanding Aptitude in Textual Entailment), a new framework for quantitative reasoning in textual entailment. We benchmark the performance of 9 published NLI models on \textsc{equate}, and find that on average, state-of-the-art methods do not achieve an absolute improvement over a majority-class baseline, suggesting that they do not implicitly learn to reason with quantities. We establish a new baseline \textsc{Q-Reas} that manipulates quantities symbolically. In comparison to the best performing NLI model, it achieves success on numerical reasoning tests (+24.2\%), but has limited verbal reasoning capabilities (-8.1\%). We hope our evaluation framework will support the development of models of quantitative reasoning in language understanding.

\end{abstract}
\section{Introduction}

Numbers play a vital role in our lives. We reason with numbers in day-to-day tasks ranging from handling currency to reading news articles to understanding sports results, elections and stock markets. As numbers are used to communicate information accurately, reasoning with them is an essential core competence in understanding natural language \cite{levinson2001pragmatics, frank2008number, dehaene2011number}. A benchmark task in natural language understanding is natural language inference (NLI)(or recognizing textual entailment (RTE)) \cite{cooper1996using, condoravdi2003entailment, bos2005recognising, dagan2006pascal}, wherein a model determines if a natural language hypothesis can be justifiably inferred from a given premise\footnote{Often, this is posed as a three-way decision where the hypothesis can be inferred to be true (entailment), false (contradiction) or cannot be determined (neutral).}. Making such inferences often necessitates reasoning about quantities. 

\begin{table}[tb]
\small
\setlength\tabcolsep{2pt}
\begin{tabular}{p{0.3 cm} p{6.9cm}}
\hline \\[-1.75ex] \multicolumn{2}{l}{\textsc{\textbf{RTE-Quant}} }\\ \hline \\[-1.75ex] 
\textbf{P}: & After the deal closes, Teva will generate  \textcolor{red}{sales of} about \textcolor{red}{\$ 7 billion} a year, the company said. \\
\textbf{H}: & Teva \textcolor{red}{earns \$ 7 billion} a year.\\
\hline \\[-1.75ex] \multicolumn{2}{l}{\textsc{\textbf{AWP-NLI}}} \\ \hline \\[-1.75ex]
\textbf{P}: & Each of farmer Cunningham's \textcolor{red}{6048 lambs} is either black or white  and there are \textcolor{red}{193 white ones.} \\
\textbf{H}: & \textcolor{red}{5855} of Farmer Cunningham's \textcolor{red}{lambs are black.}\\
\hline \\[-1.75ex] \multicolumn{2}{l}{\textsc{\textbf{NewsNLI}}} \\ \hline \\[-1.75ex]
\textbf{P}: & \textcolor{red}{Emmanuel Miller, 16}, and \textcolor{red}{Zachary Watson, 17}, are charged as adults, police said. \\
\textbf{H}: & \textcolor{red}{Two teen suspects} charged as adults.\\
\hline \\[-1.75ex] \multicolumn{2}{l}{\textsc{\textbf{RedditNLI}}} \\ \hline \\[-1.75ex]
\textbf{P}: & Oxfam says richest \textcolor{red}{one percent} to own \textcolor{red}{more than rest} by 2016. \\
\textbf{H}: & Richest \textcolor{red}{1\%} To Own \textcolor{red}{More Than Half} Worlds Wealth By 2016 Oxfam.\\
\hline
\end{tabular}
\caption{Examples from evaluation sets in EQUATE}
\label{tab:dataex}
\end{table}

Consider the following example from Table \ref{tab:dataex},\\

\noindent
P: With 99.6\% of precincts counted , Dewhurst held 48\% of the vote to 30\% for Cruz . \\
H:  Lt. Gov. David Dewhurst fails to get 50\% of primary vote.\\

 To conclude the hypothesis is inferable, a model must reason that since 99.6\% precincts are counted, even if all remaining precincts vote for Dewhurst, he would fail to get 50\% of the primary vote. Scant attention has been paid to building datasets to evaluate this reasoning ability. To address this gap, we present EQUATE (Evaluating Quantity Understanding Aptitude in Textual Entailment) (\S \ref{data:equate}), which consists of five evaluation sets, each featuring different facets of quantitative reasoning in textual entailment (Table \ref{tab:dataex}) (including verbal reasoning with quantities, basic arithmetic computation, dealing with approximations and range comparisons.).
 
 
 We evaluate the ability of existing state-of-the-art NLI models to perform quantitative reasoning (\S \ref{models:existing}), by benchmarking 9 published models on EQUATE. Our results show that most models are incapable of quantitative reasoning, instead relying on lexical cues for prediction. Additionally, we build \Qreas, a shallow semantic reasoning baseline for quantitative reasoning in NLI (\S \ref{models:ours}). \Qreas is effective on synthetic test sets which contain more quantity-based inference, but shows limited success on natural test sets which require deeper linguistic reasoning. However, the hardest cases require a complex interplay between linguistic and numerical reasoning. The EQUATE evaluation framework makes it clear where this new challenge area for textual entailment stands.

\section{Related Work}
NLI has attracted community-wide interest as a stringent test for natural language understanding  \cite{cooper1996using, fyodorov2000natural, glickman2005web, haghighi2005robust, harabagiu2006methods,romanoinvestigating,dagan2006pascal,giampiccolo2007third, zanzotto2006learning,malakasiotis2007learning, maccartney2009natural, de2009multi, dagan2010fourth, angeli2014naturalli, marelli2014sick}. Recently, the creation of large-scale datasets \cite{snli,williams2017broad,khot2018scitail} spurred the development of many neural models \cite{ parikh2016decomposable, nie-bansal:2017:RepEval, conneau-EtAl:2017:EMNLP2017,balazs-EtAl:2017:RepEval,chen-EtAl:2017:Long3, radford2018improving, devlin2018bert}.  

However, state-of-the-art models for NLI treat the task like a matching problem, which appears to work in many cases, but breaks down in others. As the field moves past current models of the matching variety to ones that embody more of the reasoning we know is part of the task, we need benchmarks that will enable us to mark progress in the field. Prior work on challenge tasks has already made headway in defining tasks for subproblems such as lexical inference with  hypernymy, co-hyponymy, antonymy \cite{glockner-shwartz-goldberg:2018:Short, naik-EtAl:2018:C18-1}. In this work, we specifically probe into quantitative reasoning.
 
 \newcite{de-marneffe-etal-2008-finding} find that in a corpus of real-life contradiction pairs collected from Wikipedia and Google News, \emph{29\%} contradictions arise from numeric discrepancies, and in the RTE-3 (Recognizing Textual Entailment) development set, numeric contradictions make up 8.8\% of contradictory pairs. \newcite{naik-EtAl:2018:C18-1} find that model inability to do numerical reasoning causes 4\% of errors made by state-of-the-art models. \newcite{sammons2010ask, clark2018knowledge} argue for a systematic knowledge-oriented approach in NLI by evaluating specific semantic analysis tasks, identifying quantitative reasoning in particular as a focus area. \newcite{bentivogli-etal-2010-building} propose creating specialized datasets, but feature only 6 examples with quantitative reasoning. Our work bridges this gap by providing a more comprehensive examination of quantitative reasoning in NLI. 
 

 

While to the best of our knowledge, prior work has not studied quantitative reasoning in NLI, \newcite{roy2017reasoning} propose a model for a related subtask called \emph{quantity entailment}, which aims to determine if a given quantity can be inferred from a sentence. In contrast, our work is concerned with general-purpose textual entailment which considers if a given \emph{sentence} can be inferred from another. Our work also relates to solving arithmetic word problems \cite{hosseini2014learning, mitra2016learning, zhou2015learn, upadhyay2016learning, huang2017learning, kushman2014learning, koncel2015parsing, roy2016solving, roy2017reasoning, ling2017program}. A key difference is that word problems focus on arithmetic reasoning, while the requirement for linguistic reasoning and world knowledge is limited as the text is concise, straightforward, and self-contained \cite{hosseini2014learning,Kushman2014LearningTA}. Our work provides a testbed that evaluates basic arithmetic reasoning while incorporating the complexity of natural language. 

Recently, \newcite{Dua2019DROPAR} also recognize the importance of quantitative reasoning for text understanding. They propose DROP, a reading comprehension dataset focused on a limited set of discrete operations such as counting, comparison, sorting and arithmetic. In contrast, \textsc{EQUATE} features diverse phenomena that occur naturally in text, including reasoning with approximation, ordinals, implicit quantities and quantifiers, requiring NLI models to reason comprehensively about the interplay between quantities and language. Additionally, through EQUATE we suggest the inclusion of controlled synthetic tests in evaluation benchmarks. Controlled tests act as basic validation of model behaviour, isolating model ability to reason about a property of interest.

\begin{table*}[tb]
\small
\resizebox{\textwidth}{!}{ 
\begin{tabular}{|p{1.08cm}|p{1.9cm}|p{0.9cm}|p{1cm}|p{2.5cm}|p{2cm}|p{3.5cm}|}
\hline
\textbf{Source} & \textbf{Test Set} & \textbf{Size} & \textbf{Classes} & \textbf{Data Source} & \textbf{Annotation Source} &  \textbf{Quantitative Phenomena }\\ \hline
& \rte & 166 & 2 & RTE2-RTE4 & Experts & Arithmetic, Ranges,  Quantifiers\\ \cline{2-7}
Natural & \qnli & 968 & 2 & CNN & Crowdworkers &   Ordinals, Quantifiers, Arithmetic, Approximation, Magnitude, Ratios\\ \cline{2-7}
& \reddit & 250& 3 & Reddit & Experts &  Range, Arithmetic, Approximation, Verbal\\ \hline
& \st & 7500 & 3 & AQuA-RAT & Automatic&  Quantifiers \\ \cline{2-7}
Synthetic & \awp & 722 & 2 & Arithmetic Word Problems & Automatic &  Arithmetic\\ 
\hline
\end{tabular}
}
\caption{An overview of test sets included in EQUATE. \reddit{} and \st{} are framed as 3-class (entailment, neutral, contradiction) while \rte{}, \qnli{} and \awp{}  are 2-class (entails=yes/no). RTE 2-4 formulate entailment as a 2-way decision. We find that few news article headlines are contradictory, thus \qnli{} is similarly framed as a 2-way decision. For algebra word problems, substituting the wrong answer in the hypothesis necessarily creates a contradiction under the event coreference assumption \cite{de-marneffe-etal-2008-finding}, thus it is framed as a 2-way decision as well.}
\label{tab:equate}
\end{table*} 
\section{Quantitative Reasoning in NLI}
\label{data:equate}

Our interpretation of ``quantitative reasoning'' draws from cognitive testing and education \cite{stafford1972hereditary,ekstrom1976manual}, which considers it ``verbal problem-solving ability''. While inextricably linked to mathematics, it is an inclusive skill involving everyday language rather than a specialized lexicon. To excel at quantitative reasoning, one must interpret quantities expressed in language, perform basic calculations, judge their accuracy, and justify quantitative claims using verbal and numeric reasoning.  These requirements show a reciprocity: NLI lends itself as a test bed for quantitative reasoning, which conversely, is important for NLI \cite{sammons2010ask, clark2018knowledge}. Motivated by this, we present the EQUATE (Evaluating Quantity Understanding Aptitude in Textual Entailment) framework.

\subsection{The EQUATE Dataset}
\textsc{EQUATE} consists of five NLI test sets featuring quantities. Three of these tests for quantitative reasoning feature language from real-world sources such as news articles and social media (\S \ref{data:rte}; \S \ref{data:qnli}; \S \ref{data:reddit}). We focus on sentences containing quantities with numerical values, and consider an entailment pair to feature quantitative reasoning if it is at least one component of the reasoning required to determine the entailment label (but not necessarily the only reasoning component). Quantitative reasoning features quantity matching, quantity comparison, quantity conversion, arithmetic, qualitative processes, ordinality and quantifiers, quantity noun and adverb resolution\footnote{Such as the quantities represented in \emph{dozen}, \emph{twice}, \emph{teenagers}.} as well as verbal reasoning with the quantity's textual context\footnote{For example, $\langle$Obama cuts tax rate to 28\%, Obama wants to cut tax rate to 28\% as part of overhaul$\rangle$.}. Appendix \ref{sec:phenom} gives some examples for these quantitative phenomena. We further filter sentence pairs which require only temporal reasoning, since specialized knowledge is needed to reason about time. These three test sets contain pairs which conflate multiple lexical and quantitative reasoning phenomena. In order to study aspects of quantitative reasoning in isolation, \textsc{EQUATE} further features two controlled synthetic tests (\S \ref{data:st}; \S \ref{data:awp}), evaluating model ability to reason with quantifiers and perform simple arithmetic. We intend that models reporting performance on any NLI dataset additionally evaluate on the EQUATE benchmark, to demonstrate competence at quantitative reasoning.

\subsection{\rte}
\label{data:rte}
This test set is constructed from the RTE sub-corpus for quantity entailment \cite{roy2017reasoning}, originally drawn from the RTE2-RTE4 datasets \cite{dagan2006pascal}. The original sub-corpus conflates temporal and quantitative reasoning. We discarded pairs requiring temporal reasoning, obtaining a set of 166 entailment pairs. 



\subsection{\qnli}
\label{data:qnli}
This test set is created from the CNN corpus \cite{hermann2015teaching} of news articles with abstractive summaries. We identify summary points with quantities, filtering out temporal expressions. For a summary point, the two most similar sentences\footnote{According to Jaccard similarity.} from the article are chosen, flipping pairs where the premise begins with a first-person pronoun (eg:$\langle$``He had nine pears'', ``Bob had nine pears''$\rangle$ becomes $\langle$``Bob had nine pears'', ``He had nine pears''$\rangle$). The top 50\% of similar pairs are retained to avoid lexical overlap bias. We crowdsource annotations for a subset of this data from Amazon Mechanical Turk. Crowdworkers\footnote{We require crowdworkers to have an approval rate of 95\% on at least 100 tasks and pass a qualification test.} are shown two sentences and asked to determine whether the second sentence is definitely true, definitely false, or not inferable given the first. We collect 5 annotations per pair, and consider pairs with lowest token overlap between premise and hypothesis and least difference in premise-hypothesis lengths when stratified by entailment label. Top 1000 samples meeting these criteria form our final set. To validate crowdsourced labels, experts are asked to annotate 100 pairs. Crowdsourced gold labels match expert gold labels in 85\% cases, while individual crowdworker labels match expert gold labels in 75.8\%.  Disagreements are manually resolved by experts and examples not featuring quantitiative reasoning are filtered, leaving a set of 968 samples.

\subsection{RedditNLI}
\label{data:reddit}
This test set is sourced from the popular social forum \textbackslash reddit\footnote{According to the Reddit User Agreement, users grant Reddit the right to make their content available to other organizations or individuals.}. Since reasoning about quantities is important in domains like finance or economics, we scrape all headlines from the posts on \textbackslash r\textbackslash economics, considering titles that contain quantities and do not have meta-forum information. Titles appearing within three days of each other are clustered by Jaccard similarity, and the top 300 pairs are extracted. After filtering out nonsensical titles, such as concatenated stock prices, we are left with 250 sentence pairs. Similar to RTE, two expert annotators label these pairs, achieving a Cohen's kappa of 0.82. Disagreements are discussed to resolve final labels.

\subsection{\st}
\label{data:st}
We include the numerical reasoning stress test from \cite{naik-EtAl:2018:C18-1} as a synthetic sanity check. The stress test consists of 7500 entailment pairs constructed from sentences in algebra word problems \cite{P17-1015}. Focusing on quantifiers, it requires models to compare entities from hypothesis to the premise while incorporating quantifiers, but does not require them to perform the computation from the original algebra word problem (eg: $\langle$``NHAI employs 100 men to build a highway of 2 km in 50 days working 8 hours a day'',``NHAI employs less than 700 men to build a highway of 2 km in 50 days working 8 hours a day''$\rangle$). 

\subsection{\awp}
\label{data:awp}
To evaluate arithmetic ability of NLI models, we repurpose data from arithmetic word problems \cite{roy2016solving}. They have the following characteristic structure. First, they establish a world and optionally update its state. Then, a question is posed about the world. This structure forms the basis of our pair creation procedure. World building and update statements form the premise. A hypothesis template is generated by identifying modal/auxiliary verbs in the question, and subsequent verbs, which we call secondary verbs. We identify the agent and conjugate the secondary verb in present tense followed by the identified unit to form the final template (for example, the algebra word problem `Gary had 73.0 dollars. He spent 55.0 dollars on a pet snake. How many dollars did Gary have left?' would generate the hypothesis template `Agent(Gary) Verb(Has) Answer(18.0) Unit(dollars) left'). For every template, the correct guess is used to create an entailed hypothesis. Contradictory hypotheses are created by randomly sampling a wrong guess 
($x \in \Z^{+}$ if correct guess is an integer, and $x \in \R^{+}$ if it is a real number) \footnote{From a uniform distribution over an interval of 10 around the correct guess (or 5 for numbers less than 5), to identify plausible wrong guesses.}. We check for grammaticality, finding only 2\% ungrammatical hypotheses, which are manually corrected leaving a set of 722 pairs.
\section{Models}
We describe the 9 NLI models\footnote{Accuracy of all models on MultiNLI closely matches original publications (numbers in appendix \ref{sec:dev}).} used in this study, as well as our new baseline. The interested reader is invited to refer to the corresponding publications for further details.


\begin{figure}[tb]
\centering
\includegraphics[scale=0.26]{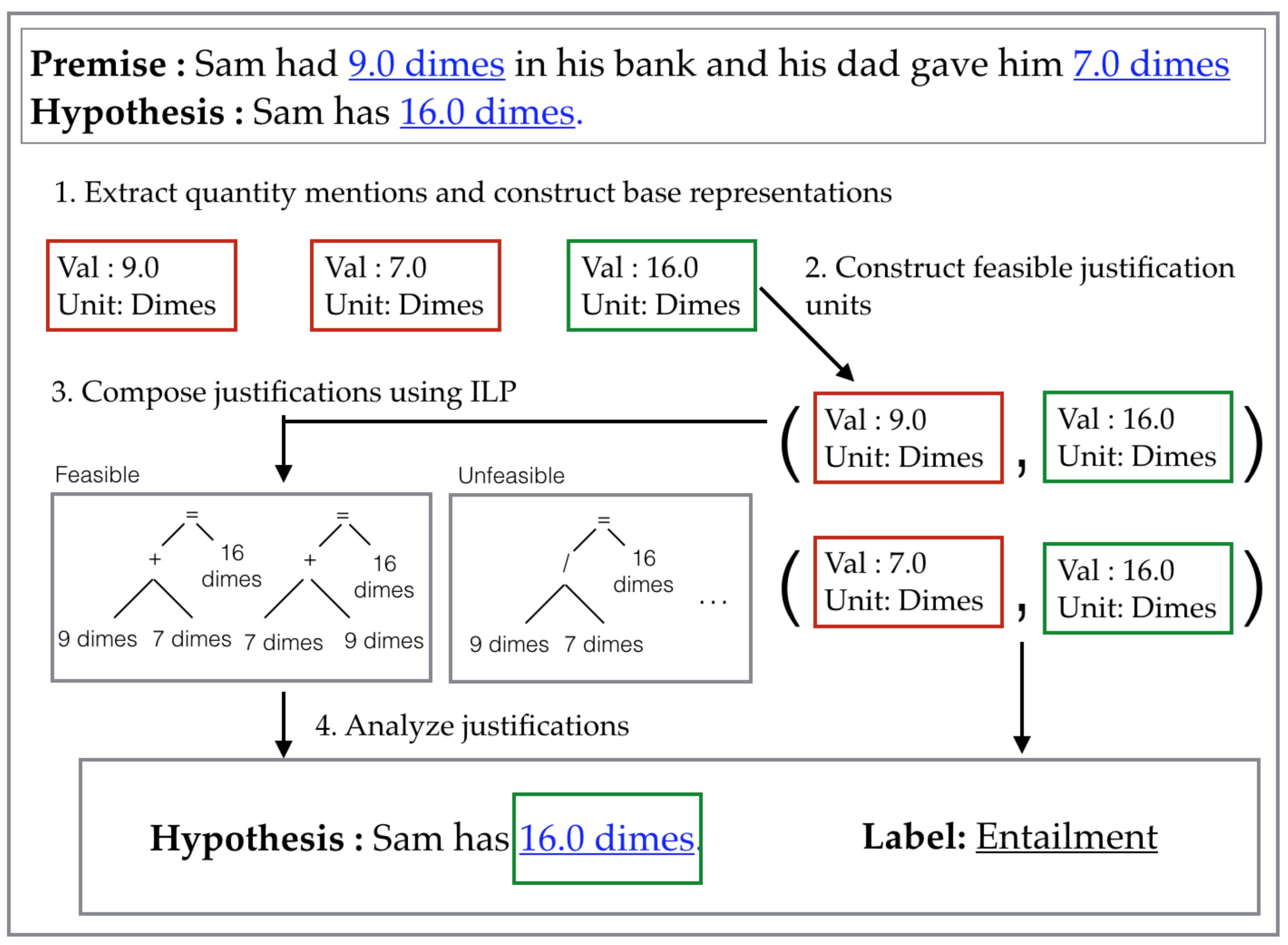}
\caption{Overview of \Qreas baseline.}
\label{fig:results_policy}
\end{figure}

\subsection{NLI Models}
\label{models:existing}
\noindent
1) \textbf{Majority Class (MAJ): } Simple baseline that always predicts the majority class in test set.\\
\noindent
2) \textbf{Hypothesis-Only (HYP): } FastText classifier \cite{mikolov2018advances} trained on only hypotheses \cite{gururangan2018annotation}. \\
\noindent
3) \textbf{ALIGN: } A bag-of-words alignment model inspired by  \newcite{maccartney2009natural}.\footnote{Model accuracy on RTE-3 test is 61.12\%, comparable to the reported average model performance in the RTE competition of 62.4\% .}\\
\noindent
4) \textbf{CBOW: } A simple bag-of-embeddings sentence representation model  \cite{williams2017broad}. \\
\noindent
5) \textbf{BiLSTM: }The simple BiLSTM model described by \newcite{williams2017broad}. \\ 
\noindent
6) \textbf{Chen (CH): }Stacked BiLSTM-RNNs with shortcut connections and character word embeddings \cite{chen-EtAl:2017:RepEval}. \\
\noindent
7) \textbf{InferSent: } A single-layer BiLSTM-RNN model with max-pooling \cite{conneau-EtAl:2017:EMNLP2017}. \\
\noindent
8) \textbf{SSEN: } Stacked BiLSTM-RNNs with shortcut connections \cite{nie-bansal:2017:RepEval}.\\
\noindent
9) \textbf{ESIM: } Sequential inference model proposed by \newcite{chen-EtAl:2017:Long3} which uses BiLSTMs with an attention mechanism. \\
\noindent
10) \textbf{OpenAI GPT:}  Transformer-based language model \cite{vaswani2017attention}, with finetuning on NLI \cite{radford2018improving}.\\
\noindent
11) \textbf{BERT:} Transformer-based language model \cite{vaswani2017attention}, with a cloze-style and next-sentence prediction objective, and finetuning on NLI \cite{devlin2018bert}.


\begin{table}[]

\small
\setlength\tabcolsep{2pt}
\resizebox{\columnwidth}{!}{
\begin{tabular}{|p{1 cm} p{6.5cm}|}
\hline \multicolumn{2}{|l|}{INPUT} \\
$P_c$ & Set of ``compatible'' single-valued premise quantities \\
$P_r$ & Set of ``compatible'' range-valued premise quantities\\
$H$ & Hypothesis quantity\\
$O$ & Operator set $\{+, -, *, /, =, \cap, \cup, \setminus, \subseteq \}$\\
$L$ & Length of equation to be generated\\
$SL$ & Symbol list ($P_c \cup P_r \cup H \cup O$)\\
$TL$ & Type list (set of types from $P_c, P_r, H$)\\
$N$ & Length of symbol list\\
$K$ & Index of first range quantity in symbol list\\
$M$ & Index of first operator in symbol list\\
\hline \multicolumn{2}{|l|}{OUTPUT} \\
$e_i$ & Index of symbol assigned to $i^{th}$ position in postfix equation\\
\hline \multicolumn{2}{|l|}{VARIABLES} \\
$x_i$ & Main ILP variable for position $i$\\
$c_i$ & Indicator variable: is $e_i$ a single value?\\
$r_i$ & Indicator variable: is $e_i$ a range?\\
$o_i$ & Indicator variable: is $e_i$ an operator?\\
$d_i$ & Stack depth of $e_i$\\
$t_i$ & Type index for $e_i$\\
\hline
\end{tabular}
}
\caption{Input, output and variable definitions for the Integer Linear Programming (ILP) framework used for quantity composition}
\label{tab:ilpvar}
\end{table}

\begin{table*}[tb]
\small
\centering
\begin{tabular}{|p{3cm}|p{11.5cm}|}
\hline \multicolumn{2}{|c|}{\textbf{Definitional Constraints}} \\ \hline
Range restriction & $x_i < K$ or $x_i = M-1$ for $i \in [0,L-1]$ if $c_i = 1$\\
 & $x_i \geq K$ and $x_i < M$ for $i \in [0,L-1]$ if $r_i = 1$\\
  & $x_i \geq M$ for $i \in [0,L-1]$ if $o_i = 1$\\
Uniqueness & $c_i + r_i + o_i = 1$ for $i \in [0,L-1]$ \\
Stack definition & $d_0 = 0$ (Stack depth initialization)\\
 & $d_i = d_{i-1} - 2o_i + 1$ for $i \in [0,L-1]$ (Stack depth update)\\
\hline \multicolumn{2}{|c|}{\textbf{Syntactic Constraints}} \\ \hline
First two operands & $c_0 + r_0 = 1$ and $c_1 + r_1 = 1$\\
Last operator & $x_{L-1} \geq N-1$ (Last operator should be one of $\{ =, \subseteq \}$)\\
Last operand & $x_{L-2} = M-1$ (Last operand should be hypothesis quantity)\\
Other operators & $x_i \leq N-2$ for $i \in [0,L-3]$ if $o_i = 1$\\
Other operands & $x_i < K$ for $i \in [0,L-3]$ if $c_i = 1$\\
 & $x_i < M$ for $i \in [0,L-3]$ if $r_i = 1$\\
Empty stack & $d_{L-1} = 0$ (Non-empty stack indicates invalid postfix expression) \\
Premise usage & $x_i \neq x_j$ for $i,j \in [0,L-1]$ if $o_i \neq 1, o_j \neq 1$ \\
\hline \multicolumn{2}{|c|}{\textbf{Operand Access}} \\ \hline
Right operand & $op2(x_i) = x_{i-1}$ for $i \in [0,L-1]$ such that $o_i = 1$\\
Left operand & $op1(x_i) = x_{l}$ for $i,l \in [0,L-1]$ where $o_i = 1$ and $l$ is the largest index such that $l \leq (i-2)$ and $d_l = d_i$\\ \hline
\end{tabular}
\caption{Mathematical validity constraint definitions for the ILP framework. Functions $op1()$ and $op2()$ return the left and right operands for an operator respectively. Variables defined in table \ref{tab:ilpvar}.}
\label{tab:ilpmath}
\end{table*}

\begin{table*}[tb]
\centering
\small
\begin{tabular}{|p{3cm}|p{11.5cm}|}
\hline \multicolumn{2}{|c|}{\textbf{Type Consistency Constraints}} \\ \hline
Type assignment & $t_i = TL[k]$ for $i \in [0,L-1]$ if $c_i + r_i = 1$ and $type(SL_i) = k$\\
Two type match & $t_i = t_a = t_b$ for $i \in [0,L-1]$ such that $o_i = 1, x_i \in \{+, -, *, /, =, \cap, \cup, \setminus, \subseteq \}, a=op1(x_i), b=op2(x_i)$ \\
One type match & $t_i \in \{ t_a, t_b\}, t_a \neq t_b$ for $i \in [0,L-1]$ such that $o_i = 1, x_i = *, a=op1(x_i), b=op2(x_i)$ \\
 & $t_i = t_a \neq t_b$ for $i \in [0,L-1]$ such that $o_i = 1, x_i = /, a=op1(x_i), b=op2(x_i)$ \\
\hline \multicolumn{2}{|c|}{\textbf{Operator Consistency Constraints}} \\ \hline
Arithmetic operators & $c_a = c_b = 1$ for $i \in [0,L-1]$ such that $o_i = 1, x_i \in \{+, -, *, /, =\}, a=op1(x_i), b=op2(x_i)$ \\ 
Range operators & $r_a = r_b = 1$ for $i \in [0,L-1]$ such that $o_i = 1, x_i \in \{ \cap, \cup, \setminus \}, a=op1(x_i), b=op2(x_i)$\\
 & $r_b = 1$ for $i \in [0,L-1]$ such that $o_i = 1, x_i = \subseteq, b=op2(x_i)$\\
\hline
\end{tabular}
\caption{Linguistic consistency constraint definitions for the ILP framework. Functions $op1()$ and $op2()$ return the left and right operands for an operator respectively. Variables defined in table \ref{tab:ilpvar}.}
\label{tab:ilpling}
\end{table*}
\subsection{\Qreas Baseline System}
\label{models:ours}

\begin{table*}[t]
\small
\centering
\resizebox{\textwidth}{!}{ 
\begin{tabular}{l l r  l r l r l r l r |l l l }
\toprule
\diagbox{M}{D}  & \multicolumn{2}{l}{RTE-Q \hfill    \raggedright{$\Delta$}}  & \multicolumn{2}{l}{NewsNLI \hfill $\Delta$} & \multicolumn{2}{l}{RedditNLI \hfill $\Delta$} & \multicolumn{2}{l}{NR ST \hfill \raggedright{$\Delta$}} & \multicolumn{2}{l}{AWPNLI \hfill $\Delta$} &  \makecell{Nat. \\Avg. $\Delta$} &  \makecell{Synth. \\Avg. $\Delta$}  & \makecell{All \\ Avg. $\Delta$} \\ \midrule
MAJ & 57.8  & \textcolor{blue}{0.0}   & 50.7  & \textcolor{blue}{0.0}   & \textbf{58.4}  & \textbf{\textcolor{blue}{0.0}} & 33.3 & \textcolor{blue}{0.0}  & 50.0  & \textcolor{blue}{0.0} & \textcolor{blue}{+0.0} & \textcolor{blue}{+0.0} & \textcolor{blue}{+0.0}\\  
HYP  & 49.4 & \textcolor{blue}{-8.4}     & 52.5  & \textcolor{blue}{+1.8} & 40.8  &  \textcolor{blue}{-17.6}  & 31.2  & \textcolor{blue}{-2.1} & 50.1  &  \textcolor{blue}{+0.1}  & \textcolor{blue}{-8.1} & \textcolor{blue}{-1.0} & \textcolor{blue}{-5.2}\\ 
ALIGN   & 62.1  & \textcolor{blue}{+4.3}   & 56.0  &  \textcolor{blue}{+5.3}  &  34.8  &  \textcolor{blue}{-23.6} & 22.6  &  \textcolor{blue}{-10.7} & 47.2 &  \textcolor{blue}{-2.8} & \textcolor{blue}{-4.7} & \textcolor{blue}{-6.8} & \textcolor{blue}{-5.5}   \\ \hline
CBOW  &47.0  &  \textcolor{blue}{-10.8}    & 61.8  &  \textcolor{blue}{+11.1}  & 42.4  & \textcolor{blue}{-16.0}  & 30.2  &  \textcolor{blue}{-3.1}   & 50.7  &  \textcolor{blue}{+0.7} & \textcolor{blue}{-5.2} & \textcolor{blue}{-1.2} & \textcolor{blue}{-3.6}  \\ 
BiLSTM  &51.2  &  \textcolor{blue}{-6.6}     &  63.3  &  \textcolor{blue}{+12.6}   & 50.8  &  \textcolor{blue}{-7.6}  & 31.2  &  \textcolor{blue}{-2.1} & 50.7  &  \textcolor{blue}{+0.7}  & \textcolor{blue}{-0.5} & \textcolor{blue}{-0.7} &  \textcolor{blue}{-0.6} \\ 
CH   &54.2  &  \textcolor{blue}{-3.6}    & 64.0  &  \textcolor{blue}{+13.3}   & 55.2  &  \textcolor{blue}{-3.2} & 30.3  &  \textcolor{blue}{-3.0}  & 50.7  &  \textcolor{blue}{+0.7}  & \textcolor{blue}{+2.2} & \textcolor{blue}{-1.2}  &  \textcolor{blue}{+0.9} \\ 
InferSent & 66.3  &  \textcolor{blue}{+8.5}   & 65.3  &  \textcolor{blue}{+14.6}   & 29.6  &  \textcolor{blue}{-28.8}  & 28.8  &  \textcolor{blue}{-4.5}  & 50.7  &  \textcolor{blue}{+0.7} & \textcolor{blue}{-1.9} & \textcolor{blue}{-1.9} &  \textcolor{blue}{-1.9}  \\ 
SSEN  &58.4  &  \textcolor{blue}{+0.6}   & 65.1  &  \textcolor{blue}{+14.4}   & 49.2  &  \textcolor{blue}{-9.2}  & 28.4  &  \textcolor{blue}{-4.9} &50.7  &   \textcolor{blue}{+0.7} & \textcolor{blue}{+1.9} & \textcolor{blue}{-2.1}  &  \textcolor{blue}{+0.3}    \\ 
ESIM   &54.8  &  \textcolor{blue}{-3.0}   & 62.0  &  \textcolor{blue}{+11.3}   & 45.6  &  \textcolor{blue}{-12.8} & 21.8  &  \textcolor{blue}{-11.5} &50.1  &  \textcolor{blue}{+0.1}  & \textcolor{blue}{-1.5} & \textcolor{blue}{-5.7}  &  \textcolor{blue}{-3.2}    \\ 
GPT                 & \textbf{68.1}   &    \textbf{\textcolor{blue}{+10.3}}           & 72.2  & \textcolor{blue}{+21.5}     & 52.4  &  \textcolor{blue}{-6.0}  & 36.4  &  \textcolor{blue}{+3.1}  & 50.0  & \textcolor{blue}{+0.0} & \textcolor{blue}{+8.6} & \textcolor{blue}{+1.6}    & \textcolor{blue}{+5.8}                                                                       \\ 
BERT         &   57.2 & \textcolor{blue}{-0.6}      & \textbf{72.8}  & \textbf{\textcolor{blue}{+22.1}}      & 49.6  &  \textcolor{blue}{-8.8}          & 36.9  &  \textcolor{blue}{+3.6}                                                                              & 42.2  & \textcolor{blue}{-7.8}    & \textcolor{blue}{+4.2} & \textcolor{blue}{-2.1} & \textcolor{blue}{+1.7}  \\
\midrule
\Qreas             & 56.6  &  \textcolor{blue}{-1.2}               & 61.1  &    \textcolor{blue}{+10.4} &  50.8  & \textcolor{blue}{-7.6}  &  \textbf{63.3}   & \textbf{\textcolor{blue}{+30}}   & \textbf{71.5}   &   \textbf{\textcolor{blue}{+21.5}} & \textcolor{blue}{+0.5} & \textcolor{blue}{+25.8} &   \textbf{\textcolor{blue}{+10.6}}  \\\bottomrule
\end{tabular}
}
\caption{Accuracies(\%) of 9 NLI Models on five tests for quantitiative reasoning in entailment. M and D represent \emph{models} and \emph{datasets} respectively. $\Delta$ captures improvement over majority-class baseline for a dataset. Column Nat.Avg. reports the average accuracy(\%) of each model across 3 evaluation sets constructed from natural sources (\rte, \qnli, \reddit), whereas Synth.Avg. reports the average accuracy(\%) on 2 synthetic evaluation sets (\st, \awp). Column Avg. represents the average accuracy(\%) of each model across all 5 evaluation sets in \textsc{equate}.  }
\label{ref:bigtable}
\end{table*}
Figure \ref{fig:results_policy} describes the \Qreas baseline for quantitative reasoning in NLI. The model manipulates quantity representations symbolically to make entailment decisions, and is intended to serve as a strong heuristic baseline for numerical reasoning on the \textsc{equate} benchmark.   This model has four stages: Quantity mentions are extracted and parsed into semantic representations called \numsets (\S\ref{ref:Qsegment}, \S\ref{ref:Qparse}); compatible \numsets are extracted (\S\ref{ref:Qprune}) and composed (\S\ref{ref:Qcompose}) to form \emph{justifications}; Justifications are analyzed to determine entailment labels (\S\ref{ref:Qjustify}). 

\subsubsection{Quantity Segmenter}
\label{ref:Qsegment}

We follow \newcite{barwise1981generalized} in defining quantities as having a number, unit, and an optional approximator. Quantity mentions are identified as least ancestor noun phrases from the constituency parse of the sentence containing cardinal numbers.

\subsubsection{Quantity Parser}
\label{ref:Qparse}
The quantity parser constructs a grounded representation for each quantity mention in the premise or hypothesis, henceforth known as a \numset\footnote{A \numset may be a composition of other \numsets .}. A \numset is a tuple (val, unit, ent, adj, loc, verb, freq, flux)\footnote{As in \cite{koncel2015parsing} S denotes all possible spans in the sentence, ${\phi}$ represents the empty span, and $S^{\phi}$=$S \cup \phi$} with:\\

\noindent
1. val $\in$ [$\R , \R$]: quantity value represented as a range\\ 
2. unit $\in S$: unit noun associated with quantity \\
3. ent $\in S^{\phi}$: entity noun associated with unit (e.g., ‘\emph{donations} worth 100\$’)\\
4. adj $\in S^{\phi}$: adjective associated with unit if any\footnote{Extracted as governing verb linked to entity by an \emph{amod} relation.},  \\
5. loc $\subseteq S^{\phi}$: location of unit (e.g.,'in the bag')\footnote{Extracted as prepositional phrase attached to the quantity and containing noun phrase.}\\
6. verb $\in S^{\phi}$: action verb associated with quantity\footnote{Extracted as governing verb linked to entity by \emph{dobj} or \emph{nsubj} relation.}. \\
7. freq $\subseteq S^{\phi}$: if quantity recurs\footnote{extracted using keywords \emph{per} and \emph{every}} (e.g, 'per hour'),\\
8. flux $\in$ \{increase to, increase from, decrease to, decrease from$\}^{\phi}$: if quantity is in a state of flux\footnote{using gazetteer: \emph{increasing}, \emph{rising}, \emph{rose}, \emph{decreasing}, \emph{falling}, \emph{fell}, \emph{drop}}.\\

To extract \textbf{values} for a quantity, we extract cardinal numbers, recording contiguity. We normalize the number\footnote{(remove ``,''s, convert written numbers to float, decide the numerical value, for example hundred fifty eight thousand is 158000, two fifty eight is 258, 374m is 3740000 etc.). If cardinal numbers are non-adjacent, we look for an explicitly mentioned range such as `to' and `between'.}. We also handle simple ratios such as quarter, half etc, and extract bounds (eg: \emph{fewer than 10 apples} is parsed to $[-\infty, 10]$ apples.)


To extract \textbf{units}, we examine tokens adjacent to cardinal numbers in the quantity mention and identify known units. If no known units are found, we assign the token in a \emph{numerical modifier} relationship with the cardinal number, else we assign the nearest noun to the cardinal number as the unit. A quantity is determined to be \textbf{approximate} if the word in an \emph{adverbial modifier} relation with the cardinal number appears in a gazetteer\footnote{roughly, approximately, about, nearly, roundabout, around, circa, almost, approaching, pushing, more or less, in the neighborhood of, in the region of, on the order of,something like, give or take (a few), near to, close to, in the ballpark of}. If approximate, range is extended to (+/-)2\% of the current value.

\subsubsection{Quantity Pruner}
The pruner constructs ``compatible'' premise-hypothesis \numset pairs. Consider the pair ``Insurgents killed \emph{7 U.S. soldiers}, set off a car bomb that killed \emph{four Iraqi policemen}'' and ``\emph{7 US soldiers} were killed, and \emph{at least 10 Iraqis} died''. Our parser extracts \numsets corresponding to \emph{``four Iraqi policemen''} and \emph{``7 US soldiers''} from premise and hypothesis respectively. But these \numsets should not be compared as they involve different units. The pruner discards such incompatible pairs. Heuristics to identify unit-compatible \numset pairs include three cases- 1) direct string match, 2) synonymy/hypernymy relations from WordNet, 3) one unit is a nationality/ job\footnote{Lists of jobs, nationalities scraped from Wikipedia.} and the other unit is synonymous with person \cite{roy2017reasoning}.




\label{ref:Qprune}
\subsubsection{Quantity Composition}
\label{ref:Qcompose}

The composition module detects whether a hypothesis \numset is justified by composing ``compatible'' premise \numsets. For example, consider the pair ``I had \textit{3 apples} but gave \textit{one} to my brother'' and ``I have \textit{two apples}''. Here, the premise \numsets{} $P_1$ (``\textit{3 apples}'') and $P_2$ (``\textit{one apple}'') must be composed to deduce that the hypothesis \numset{} $H_1$ (``\textit{2 apples}'') is justified. Our framework accomplishes this by generating postfix arithmetic equations\footnote{Note that arithmetic equations differ from algebraic equations in that they do \textit{not} contain unknown variables} from premise \numsets, that justify the hypothesis \numset\footnote{Direct comparisons are incorporated by adding ``='' as an operator.}. In this example, the expression $<P_1,P_2,-,H_1,=>$ will be generated. 

The set of possible equations is exponential in number of \numsets, making exhaustive generation intractable. But a large number of equations are invalid as they violate constraints such as unit consistency. Thus, our framework uses integer linear programming (ILP) to constrain the equation space. It is inspired by prior work on algebra word problems \cite{koncel2015parsing}, with some key differences:\\

\noindent
1. \textbf{ Arithmetic equations:} We focus on arithmetic equations instead of algebraic ones.\\
2. \textbf{Range arithmetic:} Quantitative reasoning involves ranges, which are handled by representing then as endpoint-inclusive intervals and adding the four operators ($\cup, \cap, \setminus, \subseteq$)\\ 
3. \textbf{Hypothesis quantity-driven:} We optimize an ILP model for each hypothesis \numset because a sentence pair is marked ``entailment'' iff every hypothesis quantity is justified.\\

Table \ref{tab:ilpvar} describes ILP variables. We impose the following types of constraints:\\
\textbf{1. Definitional Constraints: }Ensure that ILP variables take on valid values by constraining initialization, range, and update.\\
\textbf{2. Syntactic Constraints: }Assure syntactic validity of generated postfix expressions by limiting operator-operand ordering.\\
\textbf{3. Operand Access: }Simulate stack-based evaluation correctly by choosing correct operator-operand assignments.\\
\textbf{4. Type Consistency:} Ensure that all operations are type-compatible.\\
\textbf{5. Operator Consistency: }Force range operators to have range operands and mathematical operators to have single-valued operands.\\

Definitional, syntactic, and operand access constraints ensure mathematical validity while type and operator consistency constraints add linguistic consistency. Constraint formulations are provided in Tables \ref{tab:ilpmath} and \ref{tab:ilpling}. We limit tree depth to 3 and retrieve a maximum of 50 solutions per hypothesis \numset, then solve to determine whether the equation is mathematically correct. We discard equations that use invalid operations (division by 0) or add unnecessary complexity (multiplication/ division by 1). The remaining equations are considered plausible justifications .

\subsubsection{Global Reasoner}
\label{ref:Qjustify}
The global reasoner predicts the final entailment label as shown in Algorithm 1\footnote{MaxSimilarityClass() takes two quantities and returns a probability distribution over entailment labels based on unit match. Similarly, ValueMatch() detects whether two quantities match in value (this function can also handle ranges).}, on the assumption that every \numset in the hypothesis \emph{has} to be justified \footnote{This is a necessary but not sufficient condition for entailment. Consider the example, $\langle$`Sam believed Joan had 5 apples', `Joan had 5 apples'$\rangle$. The hypothesis quantities of 5 apples is justified but is not a sufficient condition for entailment.} for entailment. 

\begin{algorithm}[h]
\caption{PredictEntailmentLabel($P,H,C,E$)}
\label{alg:global}
\begin{algorithmic}[1]
\Input Premise quantities $P$, Hypothesis quantities $H$, Compatible pairs $C$, Equations $E$
\Output Entailment label $l \in $ \{ e, c, n \}
\If{$C$ = $\emptyset$}
\Return{$n$}
\EndIf
\State $J \leftarrow \emptyset$
\State $L \leftarrow [ ]$
\For{$q_h \in H$}
\State $J_h \leftarrow \{ q_p \mid q_p \in P, (q_p, q_h) \in C \}$
\State $J \leftarrow J \cup \{(q_h, J_h)\}$
\State $L \leftarrow L + [false]$
\EndFor
\For{$(q_h, J_h) \in J$}
\If{$J_h = \emptyset$}
\Return{$n$}
\EndIf
\For{$q_p \in J_h$}
\State  $s \leftarrow$ MaxSimilarityClass($q_p, q_h$)
\If{$s = e$}
\If{ValueMatch($q_p, q_h$)}
\State $L$[$q_h$] = $true$
\EndIf
\If{!ValueMatch($q_p, q_h$)}
\State $L$[$q_h$] = $false$ 
\EndIf
\EndIf
\If{$s = c$}
\If{ValueMatch($q_p, q_h$)}
\State $L$[$q_h$] = $c$
\EndIf
\EndIf
\EndFor
\EndFor
\For{$q_h \in H$}
\State $E_q \leftarrow \{ e_i \in E \mid hyp(e_i) = q_h \}$
\If{$E_q \neq \emptyset$}
\State $L$[$q_h$] = $true$
\EndIf
\EndFor
\If{$c \in L$}
\Return{$c$}
\EndIf
\If{count($L, true$) = len($L$)}
\Return{$e$} \\
\Return{$n$}
\EndIf
\end{algorithmic}
\end{algorithm}

\section{Results and Discussion} 

Table \ref{ref:bigtable} presents results on EQUATE. All models, except \Qreas are trained on MultiNLI. \Qreas utilizes WordNet and lists from Wikipedia. We observe that neural models, particularly OpenAI GPT excel at verbal aspects of quantitative reasoning (\rte, \qnli), whereas \Qreas excels at numerical aspects (\st, \awp).\\

 \subsection{Neural Models on \qnli{}:} 
 
To tease apart contributory effects of numerical and verbal reasoning in natural data, we experiment with \qnli. We extract all entailed pairs where a quantity appears in both premise and hypothesis, and perturb the quantity in the hypothesis generating contradictory pairs. For example, the pair $\langle$`In addition to 79 fatalities , some 170 passengers were injured.'$\rangle$ `The crash took the lives of 79 people and injured some 170', `entailment' is changed to $\langle$`In addition to 79 fatalities , some 170 passengers were injured.',  `The crash took the lives of 80 people and injured some 170', `contradiction'$\rangle$, assuming scalar implicature and event coreference. Our perturbed test set contains 218 pairs. On this set, GPT\footnote{the best-performing neural model on \textsc{EQUATE}.} achieves an accuracy of 51.18\%, as compared to 72.04\% on the unperturbed set, suggesting the model relies on verbal cues rather than numerical reasoning. In comparison, \Qreas achieves an accuracy of 98.1\% on the perturbed set, compared to 75.36\% on the unperturbed set, highlighting reliance on quantities rather than verbal information. Closer examination reveals that OpenAI switches to predicting the `neutral' category for perturbed samples instead of entailment, accounting for 42.7\% of its errors, possibly symptomatic of lexical bias issues \cite{naik-EtAl:2018:C18-1}. \\


\subsection{What Quantitative Phenomena Are Hard?}

We sample 100 errors made by \Qreas on each test in EQUATE, to identify phenomena not addressed by simple quantity comparison.  Our analysis of causes for error suggest avenues for future research:\\

\noindent
1. \textbf{Multi-step numerical-verbal reasoning:} Models do not perform well on examples requiring interleaved verbal and quantitative reasoning, especially multi-step deduction. Consider the pair $\langle$``Two people were injured in the attack'', ``Two people perpetrated the attack''$\rangle$. Quantities ``two people'' and ``two people'' are unit-compatible, but must not be compared. Another example is the \qnli{} entailment pair in Table \ref{tab:dataex}. This pair requires us to identify that 16 and 17 refer to Emmanuel and Zachary's ages (quantitative), deduce that this implies they are teenagers (verbal) and finally count them (quantitative) to get the hypothesis quantity ``two teens''. Numbers and language are intricately interleaved and developing a reasoner capable of handling such complex interplay is challenging.\\

\noindent
2. \textbf{Lexical inference:} Lack of real world knowledge causes errors in identifying quantities and valid comparisons. Errors include mapping abbreviations to correct units (``m'' to ``meters''), detecting part-whole coreference (``seats'' can be used to refer to ``buses''), and resolving hypernymy/hyponymy (``young men'' to ``boys''). \\

\noindent
3. \textbf{Inferring underspecified quantities:} Quantity attributes can be implicitly specified, requiring inference to generate a complete representation. Consider ``A mortar attack killed four people and injured 80''. A system must infer that the quantity ``80'' refers to people. On \rte, 20\% of such cases stem from zero anaphora, a hard problem in coreference resolution. \\

\noindent
4. \textbf{Arithmetic comparison limitations:} These examples require composition between incompatible quantities. For example, consider $\langle$``There were 3 birds and 6 nests'', ``There were 3 more nests than birds''$\rangle$. To correctly label this pair ``3 birds'' and ``6 nests'' must be composed.

\section{Conclusion}
In this work, we present EQUATE, an evaluation framework to estimate the ability of models to reason quantitatively in textual entailment. We observe that existing neural approaches rely heavily on the lexical matching aspect of the task to succeed rather than reasoning about quantities. We implement a strong symbolic baseline \Qreas{} that achieves success at numerical reasoning, but lacks sophisticated verbal reasoning capabilities. The \textsc{EQUATE} resource presents an opportunity for the community to develop powerful hybrid neuro-symbolic architectures, combining the strengths of neural models with specialized reasoners such as \Qreas{}. We hope our insights lead to the development of models that can more precisely reason about the important, frequent, but understudied, phenomena of quantities in natural language.

\section*{Acknowledgments}
This research was supported in part by grants from the National Science Foundation Secure and Trustworthy Computing program (CNS-1330596, CNS-15-13957, CNS-1801316, CNS-1914486) and a DARPA Brandeis grant (FA8750-15-2-0277). The views and conclusions contained herein are those of the authors and should not be interpreted as necessarily representing the official policies or endorsements, either expressed or implied, of the NSF, DARPA, or the US Government.  The author Naik was supported by a fellowship from the Center of Machine Learning and Health at Carnegie Mellon University.  The authors would like to thank Graham Neubig, Mohit Bansal and Dongyeop Kang for helpful discussion regarding this work, and Shruti Rijhwani and Siddharth Dalmia for reviews while drafting this paper. The authors are also grateful to Lisa Carey Lohmueller and Xinru Yan for volunteering their time for pilot studies.

\bibliography{conll-2019}
\bibliographystyle{acl_natbib}

\appendix
\section*{Appendix}

\section{Baseline performance on MultiNLI-Dev Matched}
\label{sec:dev}

\begin{table}[h]
\begin{tabular}{|l|l|}
\hline
 \textbf{Model }                 & \textbf{MultiNLI Dev} \\ \hline
Hyp Only           & 53.18\%      \\ \hline
ALIGN              & 45.0\%       \\ \hline
CBOW               & 63.5\%       \\ \hline
BiLSTM             & 70.2\%       \\ \hline
Chen               & 73.7\%       \\ \hline
NB                 & 74.2\%       \\ \hline
InferSent          & 70.3\%       \\ \hline
ESIM               & 76.2\%       \\ \hline
OpenAI Transformer & 81.35\%      \\ \hline
BERT & 83.8\%      \\ \hline
\end{tabular}
\caption{Performance of all baseline models used in the paper on the matched subset of MultiNLI-Dev}
\label{tab:mnlidev}
\end{table}

Table \ref{tab:mnlidev} presents classification accuracies of all baseline models used in this work on the matched subset of MultiNLI-Dev. These scores are very close to the numbers reported by the original publications, affirming the correctness of our baseline setup.

\section{Examples of quantitative phenomena present in EQUATE}
Table \ref{tab:pheno} presents some examples from EQUATE which demonstrate interesting quantitative phenomena that must be understood to label the pair correctly.
\label{sec:phenom}
\begin{table*}[tb]
    \centering
    
    \begin{tabular}{p{2.1cm} p{13cm}}
    \hline \textbf{Phenomenon} & \textbf{Example} \\ \hline
       Arithmetic & \makecell[l]{\textbf{P:} Sharper faces charges in Arizona and California\\\textbf{H:} Sharper has been charged in two states }\\ \hline
       Ranges & \makecell[l]{\textbf{P:} Between 20 and 30 people were trapped in the casino \\ \textbf{H:} Upto 30 people thought trapped in casino}\\ \hline
       Quantifiers & \makecell[l]{\textbf{P:} Poll: Obama over 50\% in Florida\\ \textbf{H:} New poll shows Obama ahead in Florida}\\ \hline
       Ordinals & \makecell[l]{\textbf{P:} Second-placed Nancy celebrated their 40th anniversary with a win \\ \textbf{H:} Nancy stay second with a win}\\ \hline
       Approximation & \makecell[l]{\textbf{P:} Rwanda has dispatched 1917 soldiers \\ \textbf{H:} Rwanda has dispatched some 1900 soldiers}\\ \hline
       Ratios & \makecell[l]{\textbf{P:} Londoners had the highest incidence of E. Coli bacteria (25\%) \\ \textbf{H:} 1 in 4 Londoners have E. Coli bacteria}\\ \hline
       Comparison & \makecell[l]{\textbf{P:} Treacherous currents took four lives on the Alabama Gulf coast\\ \textbf{H:} Rip currents kill four in Alabama}\\ \hline
       Conversion & \makecell[l]{\textbf{P:} If the abuser has access to a gun, it increases chances of death by 500\% \\ \textbf{H:} Victim five times more likely to die if abuser is armed}\\ \hline
       Numeration & \makecell[l]{\textbf{P:} Eight suspects were arrested\\\textbf{H:} 8 suspects have been arrested}\\ \hline
       Implicit Quantities & \makecell[l]{\textbf{P:} The boat capsized two more times\\\textbf{H:} His sailboat capsized three times}\\ \hline
    \end{tabular}
    \caption{Examples of quantitative phenomena present in EQUATE}
    \label{tab:pheno}
\end{table*}{}


\end{document}